\documentclass[11pt]{article}

\usepackage{acl}

\usepackage{times}
\usepackage{latexsym}

\usepackage[T1]{fontenc}

\usepackage[utf8]{inputenc}

\usepackage{microtype}

\usepackage{inconsolata}

\usepackage{graphicx}

\usepackage[utf8]{inputenc} 
\usepackage[T1]{fontenc}    

\usepackage{url}            
\usepackage{booktabs}       
\usepackage{amsfonts}       
\usepackage{nicefrac}       
\usepackage{microtype}      
\usepackage{xcolor}         

\usepackage{graphicx, subcaption}
\usepackage{color}
\usepackage{enumitem}
\usepackage{makecell}
\usepackage{multirow}
\usepackage[most]{tcolorbox}
\usepackage{wrapfig}

\newcommand{\ie}{\emph{i.e., }}
\newcommand{\eg}{\emph{e.g., }}

\newcommand{\cf}{\emph{cf. }}

\title{Chronos: Learning Temporal Dynamics of Reasoning Chains \\ for Test-Time Scaling}

\author{
Kai Zhang\thanks{Equal Contributions.},
Jiayi Liao\footnotemark[1],
Chengpeng Li,
Ziyuan Xie,
Sihang Li\thanks{Corresponding Author.},
Xiang Wang\footnotemark[2]
\\[3pt]
University of Science and Technology of China
\\
\{kaizhang99, xzy101\}@mail.ustc.edu.cn, \\
\{joyliao7777, sihang0520, xiangwang1223\}@gmail.com\\
\vspace{-4ex}
}

\begin{document}
\maketitle
\begin{abstract}
Test-Time Scaling (TTS) has emerged as an effective paradigm for improving the reasoning performance of large language models (LLMs). 
However, existing methods --- most notably majority voting and heuristic token-level scoring --- treat reasoning traces or tokens equally, thereby being susceptible to substantial variations in trajectory quality and localized logical failures. 
In this work, we introduce \textbf{Chronos}, a lightweight and plug-and-play chronological reasoning scorer that models each trajectory as a time series. 
Specifically, Chronos learns to capture trajectory features of token probabilities, assigns quality scores accordingly, and employs a weighted voting mechanism.
Extensive evaluations on both in-domain and out-of-domain benchmarks demonstrate that Chronos consistently delivers substantial gains across a variety of models, with negligible computational overhead.
Notably, Chronos@128 achieves relative improvements of 34.21\% over Pass@1 and 22.70\% over Maj@128 on HMMT25 using Qwen3-4B-Thinking-2507, highlighting its effectiveness.
\end{abstract}
\section{Introduction}
Test-Time Scaling (TTS) \cite{inference_scaling, llmmonkeys} has emerged as a powerful paradigm that complements model training, substantially improving the performance of large language models (LLMs) \cite{gpt5,Qwen,Gemini,Deepseek} across a wide range of complex reasoning tasks \cite{QA23, codetree, mathllm}.
A key driver of these gains is the use of parallel multi-sample aggregation strategies, which generate multiple independent reasoning trajectories and combine them to infer a consensus solution.

\begin{figure}[t]
    \centering
    \begin{subfigure}[b]{\linewidth}
        \includegraphics[width=\linewidth]{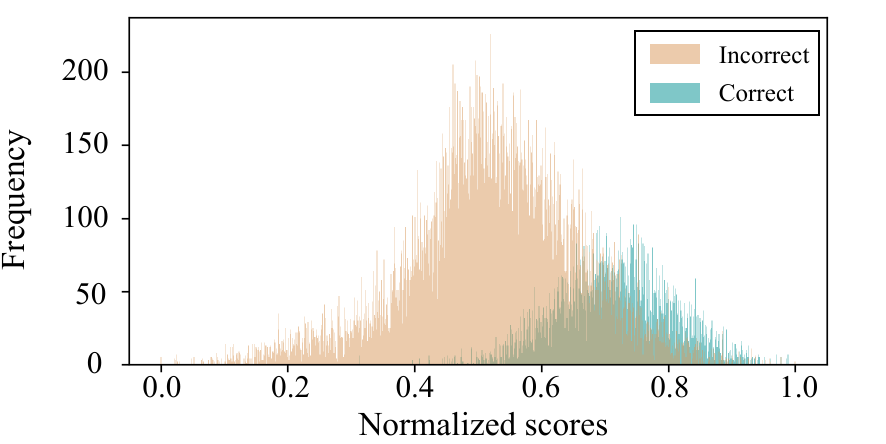}
        \caption{Tail Confidence}
    \end{subfigure}
    \begin{subfigure}[b]{\linewidth}
        \includegraphics[width=\linewidth]{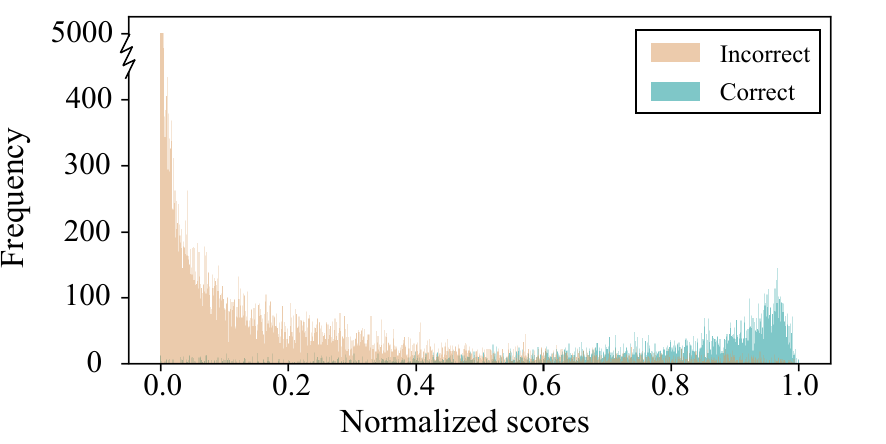}
        \caption{Chronos}
    \end{subfigure}
    \vspace{-20pt}
    \caption{Distribution of scores for correct and incorrect trajectories on AIME25. (a) Tail Confidence and (b) Chronos. All experiments use the DeepSeek-1.5B with 128 samples for 16 repeats per question.}
    \vspace{-16pt}
    \label{teaser-fig}
\end{figure}

Early self-consistency methods aggregate final answers via majority voting \cite{Self-Consistency}. 
While effective, this approach suffers from a fundamental limitation: it treats all reasoning trajectories as equally reliable, ignoring substantial variations in their quality.
As a result, erroneous or low-quality traces can disproportionately might influence the aggregation, leading to suboptimal predictions.
This limitation raises a critical research question: how can the quality of reasoning trajectories be accurately estimated?

Recent work \cite{Uncertainty, Self-Certainty, deepthink} has leveraged token-level distributional statistics to assess the quality of reasoning trajectories.
However, these methods largely depend on predefined heuristics --- such as token-level uncertainty \cite{Uncertainty} or confidence scores \cite{deepthink} --- and aggregate them into a single trace quality estimate via uniform mean pooling.
This design implicitly assumes that all tokens are unordered and contribute equally to the logical validity of a reasoning trajectory.
In practice, such homogenization can mask critical failures at intermediate steps, impairing the detection of localized reasoning errors and substantially weakening discriminative power. 
As shown in \ref{teaser-fig}, although tail confidence scores \cite{deepthink} can distinguish between correct and incorrect reasoning trajectories, there are still significant overlaps that incur ambiguity, resulting in suboptimal estimation of trajectory quality.

To overcome the limitations aforementioned, we introduce \textbf{Chronos} (\textbf{Chrono}logical Reasoning \textbf{S}corer), a versatile, plug-and-play module for high-fidelity trajectory aggregation.
In contrast to heuristic scorers \cite{deepthink} that model reasoning traces as unordered collections of token-level statistics, we posit that the validity of a reasoning chain is inherently a chronological process.
Chronos maps each token to a discrete timestamp and explicitly models the directional progression of reasoning, enabling the scorer to capture sequential dependencies.
We implement Chronos as a lightweight learned temporal scoring model that treats the inference trajectory as a time series.
The model ingests sequences of token-level probabilities --- reflecting the model’s internal signals --- and produces a scalar quality score for each reasoning trajectory.
Specifically, Chronos employs a multi-scale convolutional architecture to process the internal temporal signal. By utilizing parallel convolutional filters with varying kernel lengths, the model simultaneously captures local fluctuations and global dependencies within the reasoning process. These multi-scale blocks are stacked within a deep residual framework to capture complex temporal patterns, enabling high-fidelity estimation of trajectory quality.

We evaluate Chronos by training it on AIME (2000–2023) and conducting both in-domain and out-of-domain evaluations across multiple reasoning benchmarks --- AIME25, HMMT25, and GPQA-Diamond --- and model scales, including DeepSeek-1.5B, Qwen3-4B, and DeepSeek-8B.
Across all settings, Chronos consistently yields substantial gains in TTS performance while introducing negligible computational overhead.
Notably, when paired with Qwen3-4B-Thinking, Chronos@128 achieves 74.38\% accuracy on HMMT25 --- effectively saturating the benchmark --- compared to 55.42\% for Pass@1 and 60.62\% for Maj@128 (majority voting), with only a 0.0005\% increase in inference FLOPs.

\section{Related Work}

\paragraph{Scaling Test-Time Compute.} 
Recent advancements in LLMs \cite{inference_scaling, llmmonkeys, Scaling-Test-Time} have emphasized scaling test-time compute along two complementary paradigms: sequential refinement and parallel exploration. 
Sequential scaling aims to deepen reasoning by iteratively refining solutions \cite{Self-Refine} or by encouraging longer Chain-of-Thought (CoT) \cite{cot} traces through reinforcement learning \cite{o1,Deepseek,Qwen}. 
In contrast, parallel scaling broadens the search space via multi-trajectory sampling \cite{Self-Consistency, start2025} or structured tree search methods \cite{ToT, MCTS}. 
Despite their effectiveness, a key bottleneck lies in synthesizing the parallel outputs.
Existing approaches largely rely on predefined heuristics \cite{Mirror-Consistency,Semantic-Self-Consistency,Self-Certainty,deepthink}, which struggle to distinguish correct solutions from hallucinations, as they either treat all reasoning steps uniformly or apply rule-based aggregation (\eg uniform mean pooling), disregarding the temporal dynamics of the reasoning process. 
To address this limitation, we model reasoning as a chronological process and leverages internal signals to accurately estimate trajectory quality.

\paragraph{Time Series Modeling.} Modeling sequential data is central to various domains, evolving significantly from traditional distance-based heuristics \cite{DTW,sax} to modern deep learning approaches. To capture complex non-linear dependencies and long-term correlations, researchers have explored diverse architectures, including MLP-based mixers \cite{N-BEATS,DLinear}, recurrent networks \cite{LSTNet,DeepAR}, temporal convolutional networks (TCNs) \cite{tcn,SCINet}, and Transformers \cite{Log-Sparse-Attention,Informer}. Drawing inspiration from this progress, we reframe the evaluation of LLM reasoning chains as a time-series classification task. We treat token probability sequences as temporal signals and propose a lightweight aggregator based on the InceptionTime architecture \cite{inceptionnet,inceptiontime}. By leveraging multi-scale convolutional filters, our method effectively captures the chronological evolution of the reasoning process, achieving high-fidelity trajectory scoring with negligible computational overhead.

\section{Methodology}
\begin{figure*}[t]
    \centering
    \includegraphics[width=\linewidth]{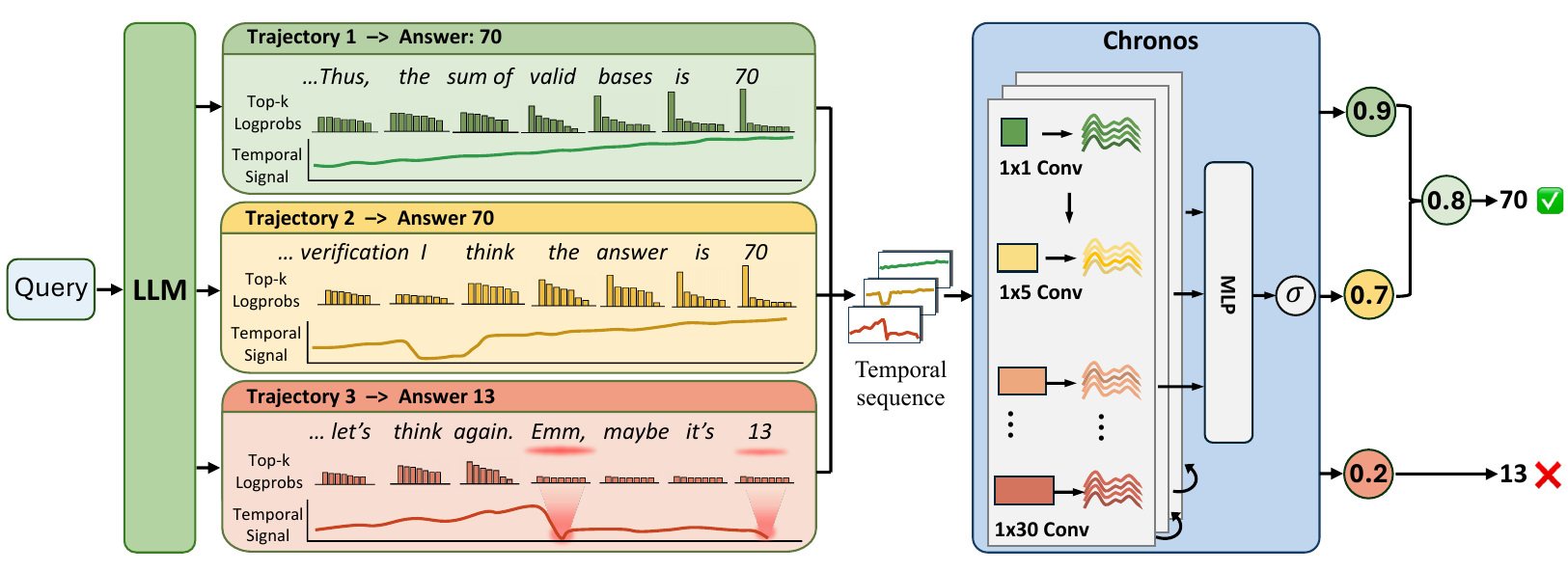}
    \caption{Framework of \textbf{Chronos}. It consists of three stages: (1) \textbf{Multi-trajectory Sampling}: Given an input query, we sample multiple independent inference trajectories and extract their token-level probability distributions, which are treated as temporal signals. (2) \textbf{Chronological Reasoning Scoring}: Chronos processes these temporal signals using multi-scale convolutions to explicitly capture sequential dependencies, producing a scalar quality score for each trajectory. (3) \textbf{Weighted Majority Voting}: The predicted trajectory scores are used to weight candidate answers, which are then aggregated to determine the final output.}
    \label{fig:framework}
 
\end{figure*}

In this section, we introduce \textbf{Chronos}, a lightweight and plug-and-play chronological reasoning scorer that models reasoning trajectories as temporal processes (\cf Figure \ref{fig:framework}).
We first describe the token-level statistics that serve as the temporal signals for Chronos in Section \ref{sec:token-level}.
We then present the Chronos architecture in Section \ref{sec:chronos}, detailing how it captures sequential dependencies in the reasoning process.
Finally, in Section \ref{sec:maj},we introduce the weighted majority voting scheme that leverages the predicted scores for aggregation.

\subsection{Token-level Statistics}
\label{sec:token-level}
Transformer-based \cite{attention} LLMs generate output token sequences $y=(y_1,...,y_m)$ autoregressively, conditioned on an input sequence $x=(x_1,...,x_n)$.
At each decoding step $t$, the model maps the preceding context to a logit vector $\textbf{v}_t\in \mathbb{R}^{|\mathcal{V}|}$, where $\mathcal{V}$ denotes the vocabulary.
These logits are transformed via the softmax operator into a probability distribution over the next token: $y_t$:
\begin{equation}
    P_t(\cdot | x, y_{<t})=\text{Softmax}(\textbf{v}_t) \in [0,1]^{|\mathcal{V}|}.
\end{equation}
This token-level distribution reflects the model’s confidence in predicting the next token.

Unlike prior approaches that collapse token-level signals into a single scalar via uniform pooling, Chronos preserves fine-grained sequential dependencies by retaining token-level probabilities of the reasoning trajectory as a chronological sequence $\mathbf{s}$:
\begin{align}
    \mathbf{s} &= (s_1, s_2, \dots, s_L), \\
    s_t &= - \frac{1}{k} \sum_{i=1}^{k} \log P_t(i \mid x, y_{<t})
\end{align}
where $s_t$ denotes the negative mean log-probability of the top-$k$ candidate tokens at decoding step $t$. 
High values of $s_t$ corresponds to more peaked distributions and greater model confidence, while low values indicate increased uncertainty.
This chronological representation preserves the temporal structure of the reasoning process, enabling Chronos to capture localized confidence fluctuations that are obscured by uniform pooling strategies.

\subsection{Model Architecture}
\label{sec:chronos}
Inspired by prior work in temporal signal processing \cite{inceptionnet,inceptiontime}, which shows that lightweight convolutional architectures can effectively capture temporal patterns, Chronos models the dynamic evolution of token-level statistics throughout the reasoning process, enabling accurate estimation of trajectory quality.

\subsubsection{Multi-scale Feature Extraction}
\label{conv_module}
The core is the multi-scale convolutional block, designed to process the temporal signal $\mathbf{s}$ at varying windows. 
Motivated by the observation that the critical reasoning steps and final answer are typically concentrated in the last stages of the trajectory, Chronos focuses exclusively on the final $L_{tail}$ tokens. 
Formally, we define the input as the temporal sequence $\mathbf{s} \in \mathbb{R}^{1\times L_{tail}}$, where the channel dimension is 1 and the sequence length is $L_{tail}$.
To augment model capacity, we first apply $N_{Proj}$ filters with the shape of $1 \times 1$ to project it into $\mathbf{z}$:
\begin{equation}
    \mathbf{z} = \text{Conv}(\mathbf{s},1,N_{Proj}) \in \mathbb{R}^{N_{Proj}\times L_{tail}},
\end{equation}
where $\text{Conv}(\cdot)$ represents the convolution operation utilizing $N_{Proj}$ filters with the shape of $1 \times 1$.
The expanded signal is then processed by parallel convolutional filters with varying kernel lengths $l$:
\begin{equation}
    \mathbf{h}_l = \text{ReLU}(\text{Conv}(\mathbf{z},l,N_{Conv}))\in \mathbb{R}^{N_{Conv} \times L_{tail}}.
\end{equation}
This process involves the convolution operation utilizing $N_{Conv}$ filters with the shape of $1 \times l$.
Particularly, shorter filters are sensitive to local fluctuations and immediate inconsistencies in the reasoning steps, while longer filters capture broader trends and long-term dependencies within the reasoning trajectory.
The outputs of these parallel convolutions, including $\mathbf{z}$ --- spanning local details and global context --- are concatenated along the channel dimension to form a rich, multi-scale representation $\mathbf{o} \in \mathbb{R}^{(N_{Proj} + k \cdot N_{Conv}) \times L_{tail}}$:
\begin{equation}
    \mathbf{o} = \text{Concat}(\mathbf{z}, \mathbf{h}_{l_1}, \mathbf{h}_{l_2}, \dots, \mathbf{h}_{l_k}),
\end{equation}
where $k$ is the number of distinct kernel lengths, and $\mathbf{h}_{l_i}$ denotes the output feature map generated by the convolution with kernel length $l_i$. 
\subsubsection{Deep Residual Architecture}
To facilitate the training of a deep network capable of capturing long-context temporal patterns, we stack $N_{Blk}$ multi-scale convolutional blocks in a sequential manner, integrated with residual connections for each block to mitigate the vanishing gradient problem.
Formally, let $\mathbf{M}_i(\cdot)$ denote the $i$-th convolutional block and $\mathbf{o}_i$ represent the output of $\mathbf{M}_i$, The forward process is defined as:
\begin{equation}
    \mathbf{o}_i = \mathbf{M}_{i}(\mathbf{o}_{i-1}), \; \hat{y}=\sigma(\text{MLP}(\sum_{i=0}^{N_{Blk}}\mathbf{o}_{i})),
\end{equation}
where $\mathbf{o}_0=\mathbf{z}$ is the expanded input embedding, $\sigma(\cdot)$ is the sigmoid function, and $\hat{y}$ denotes the final predicted score.
For a given trajectory $\tau_{i}$ from dataset $\{\tau_i\}_{i=1}^N$, we assign a ground-truth label $y_i \in \{0, 1\}$ based on the correctness of the final answer, where $y_i=1$ indicates a correct prediction and $y_i=0$ denotes an error.
The model is trained to minimize the Binary Cross-Entropy (BCE) loss between the predicted score $\hat{y}_i$ and the label $y_i$:
\begin{equation}
    \mathcal{L} = - \sum_{i=1}^{N} \left[ y_i \cdot \log(\hat{y}_i) + (1 - y_i) \cdot \log(1 - \hat{y}_i) \right].
\end{equation} 

In practice, an ensemble strategy is adopted wherein multiple well-trained models independently evaluate the trajectory, with their average serving as the final score. As shown in Figure \ref{teaser-fig}, by modeling reasoning traces as time series, Chronos effectively reduce the overlapping ambiguity among trajectories and substantially improves the accuracy of trajectory quality prediction.

\subsection{Score-Weighted Majority Voting}
\label{sec:maj}
To further boost aggregation performance, we implement a score-based filtering \cite{deepthink} alongside weighted majority voting. 
Formally, given a question $Q$, we sample $N_{smp}$ reasoning trajectories $\{\tau_i\}_{i=1}^{N_{smp}}$ and predict their corresponding scores $\{s_i\}_{i=1}^{N_{smp}}$. We sort the trajectories in descending order of $s_i$ and identify the subset of indices $\mathcal{I}_{top}$:
\begin{equation}
    \mathcal{I}_{top} = \left\{ i \mid \text{rank}(\hat{y}_i) \le \lfloor \eta \cdot N_{sam} \rfloor \right\},
\end{equation}
where $\eta$ is the retention ratio.
The final answer $\hat{a}$ is then determined by a weighted majority vote over this filtered subset:
\begin{equation}
    \hat{a} = \operatorname*{arg\,max}_{a \in \mathcal{A}} \sum_{i \in \mathcal{I}_{top}} \hat{y}_i \cdot \mathbb{I}(a_i = a),
\label{equ:maj}
\end{equation}
where $\mathcal{A}$ denotes the set of unique candidate answers for $Q$, $a_i$ represents the final answer extracted from trajectory $\tau_i$, and $\mathbb{I}(\cdot)$ is the indicator function, which equals 1 if $a_i$ matches candidate $a$, and 0 otherwise.
This approach prioritizes high-quality reasoning traces by retaining only the top $\eta$ of trajectories based on their predicted scores, thereby ensuring the final consensus relies exclusively on the most reliable traces.

\section{Experiments}
We conducted extensive experiments to demonstrate the effectiveness of Chronos. 
Specifically, our experiments aim to address the following research questions: 
\begin{itemize}[leftmargin=*,nosep]
    \item \textbf{RQ1:} How does Chronos perform compared to existing TTS methods?
    \item \textbf{RQ2:} What is the computational overhead introduced by Chronos, and how does accuracy evolve with the scaling of test-time computation?
    \item \textbf{RQ3:} How well does Chronos generalize when applied to different sampling models?
\end{itemize}
\subsection{Experimental Setup}
\paragraph{Models.}
We evaluate Chronos across multiple model scales, including: \textbf{DeepSeek-1.5B}\footnote{DeepSeek-1.5B: \url{https://huggingface.co/deepseek-ai/DeepSeek-R1-Distill-Qwen-1.5B}.}, \textbf{Qwen3-4B}\footnote{Qwen3-4B: \url{https://huggingface.co/Qwen/Qwen3-4B-Thinking-2507}.} and \textbf{DeepSeek-8B}\footnote{DeepSeek-8B: \url{https://huggingface.co/deepseek-ai/DeepSeek-R1-0528-Qwen3-8B}.}.
We selected these models due to their proven efficacy in reasoning tasks.
Complete generation settings and prompting templates are provided in Appendix \ref{app:exp-settings}. 

\paragraph{Datasets.} 
To prevent data leakage, we train Chronos exclusively on the AIME archive spanning 2000 – 2023, ensuring that the training data is comparable in difficulty to the evaluation benchmarks.
We evaluate Chronos on AIME25, HMMT25 (Feb), and GPQA-Diamond, which are widely used benchmarks for assessing the reasoning capabilities of frontier LLMs.
Notably, AIME and HMMT consist of competitive mathematics problems, whereas GPQA-Diamond comprises scientific reasoning questions, thereby enabling a rigorous assessment of Chronos’s cross-domain generalization.
Additional dataset details are provided in Appendix \ref{app:datasets}.

\paragraph{Evaluation.} 
We employ Pass@1 and majority voting as foundational evaluation metrics.
Pass@1 measures the accuracy of a single reasoning chain, while Maj@K aggregates K parallel reasoning paths by selecting the most frequent answer, treating all trajectories as equally reliable.
In addition, we include DeepConf \cite{deepthink} as a representative weighted voting baseline.
DeepConf assigns trajectory weights based on statistical confidence estimated from token-level probability distributions, in contrast to the uniform weighting used in standard majority voting.
Following the official implementation, we report the best performance achieved across mean, bottom, and tail confidence variants with varying top-$\eta$ filtering settings.

\paragraph{Implementation Details.} 
For the AIME (2000–2023) dataset, we sample 32 complete reasoning trajectories for each question and randomly partition them into training, validation, and test sets following an 8:1:1 ratio.
In our experimental setup, we fix the input sequence length to $L_{tail}=2048$, set the number of multi-scale convolutional blocks to $N_{Blk}=3$, and the retention ratio $\eta=0.1$. 
We perform a hyperparameter search for the remaining components: the projection dimension $N_{Proj}$ is selected from \{8, 16\}, the number of filters $N_{Conv}$ is searched over \{4, 8, 16\}, the number of distinct kernel lengths $k=3$, and the set of multi-scale convolution filter lengths $l$ is chosen from \{\{10, 20, 40\}, \{20, 40, 80\}, \{40, 80, 160\}\}. 

After training Chronos, we select the hyperparameter configuration that yields the highest AUC on the test set to serve as the scorer for the final evaluation.
During the evaluation phase, we first generate a candidate pool of 512 complete trajectories per question. For each experiment, we subsample 128 trajectories per question from this pool to apply the score-weighted majority voting method. 

\begin{table*}[t]
    \centering
    \caption{Performance comparison. Accuracy (\%) is reported. All experiments are repeated 16 times. The top two results are highlighted in \textbf{bold} and \underline{underlined}.}
    \vspace{-6pt}
    \label{tab:performance}
    \renewcommand{\arraystretch}{1.1}
    \resizebox{0.8\textwidth}{!}{ 
        \begin{tabular}{llcccc}
            \toprule
            Model & Dataset & \textbf{Pass@1} & \textbf{Maj@128} & \textbf{DeepConf@128} & \textbf{Chronos@128} \\
            \midrule
            \multirow{4}{*}{DeepSeek-1.5B} 
            & AIME25 & 24.58 & 37.50 & \underline{40.21} & \textbf{44.17} \\
            & HMMT25 & 12.08 & 16.67 & \underline{25.21} & \textbf{27.92} \\
            & GPQA-D & 21.67 & 33.96 & \underline{36.88} & \textbf{39.17} \\
            & \textit{Average} & 19.44 & 29.38 & \underline{34.10} & \textbf{37.09} \\ 
            \midrule
            \multirow{4}{*}{Qwen3-4B}       
            & AIME25 & 78.67 & \textbf{86.67} & 85.83 & \textbf{86.67} \\
            & HMMT25 & 55.42 & 60.62 & \underline{67.08} & \textbf{74.38} \\
            & GPQA-D & 59.79 & \underline{70.83} & 68.96 & \textbf{71.25} \\
            & \textit{Average} & 63.96 & 72.71 & \underline{73.96} & \textbf{77.43} \\ 
            \midrule
            \multirow{4}{*}{DeepSeek-8B}  
            & AIME25 & 75.21 & 83.96 & \underline{86.46} & \textbf{88.75} \\
            & HMMT25 & 60.42 & 68.75 & \underline{73.13} & \textbf{76.04} \\
            & GPQA-D & 59.17 & 67.92 & \underline{70.21} & \textbf{71.25} \\
            & \textit{Average} & 64.93 & 73.54 & \underline{76.60} & \textbf{78.68} \\ 
            \bottomrule
        \end{tabular}
    }
\vspace{-8pt}
\end{table*}

\subsection{Main Results (RQ1)}

Table \ref{tab:performance} summarizes the performance of Chronos in comparison with Pass@1, Majority Voting, and the confidence-weighted baseline DeepConf across three models and three benchmarks.

We highlight the key observations below:

\paragraph{Superiority over Majority Voting.} As shown in Table \ref{tab:performance}, Chronos consistently surpasses the standard Majority Voting (Maj@128) across all experimental settings. While majority voting effectively reduces variance compared to Pass@1, it treats all reasoning paths as equally valid, often failing when correct answers are in the minority. Chronos addresses this by assigning quality-aware scores. For instance, on the DeepSeek-1.5B model, Chronos improves the average accuracy from 29.38\% (Maj@128) to 37.09\%, a substantial absolute gain of 7.71\%. Similarly, on the larger DeepSeek-8B model, Chronos pushes the average accuracy from 73.54\% to 78.68\%. These results demonstrate that identifying and weighting high-quality trajectories is significantly more effective than simple frequency-based aggregation. 

\paragraph{Advantage over Uniformly Pooled Token-Level Statistics.} 
Chronos consistently outperforms DeepConf@128 across all nine model-dataset combinations. 
DeepConf aggregates token-level statistics via uniform mean pooling, treating the reasoning process as a collection of token probabilities.
In contrast, Chronos models each reasoning trajectory as a temporal process, and evaluates correctness by exploiting internal temporal dynamics. 
The consistent performance gap --- for example, an improvement from 34.10\% (DeepConf) to 37.09\% (Chronos) on DeepSeek-1.5B --- supports our hypothesis: the temporal evolution of logical consistency provides crucial discriminative signals for distinguishing correct reasoning from hallucinations, which uniform pooling fails to capture.

\paragraph{Remarkable Gains on Challenging Benchmarks.}
The performance gains are particularly pronounced on complex reasoning tasks. 
On the HMMT25 benchmark using Qwen3-4B, Chronos achieves a striking accuracy of 74.38\%, compared to 60.62\% for Maj@128 and 67.08\% for DeepConf. 
This 13.76\% absolute improvement over majority voting highlights Chronos's ability to identify correct solutions even when they are statistically underrepresented in the sample pool. 
Furthermore, Chronos demonstrates strong generalization on the out-of-domain GPQA-D benchmark, consistently outperforming baselines across all model scales, proving its robustness beyond mathematical domains. 

\subsection{Scaling Experiments (RQ2)}

\begin{figure*}[t]
    \centering
    \begin{subfigure}[b]{\linewidth}
        \includegraphics[width=\linewidth]{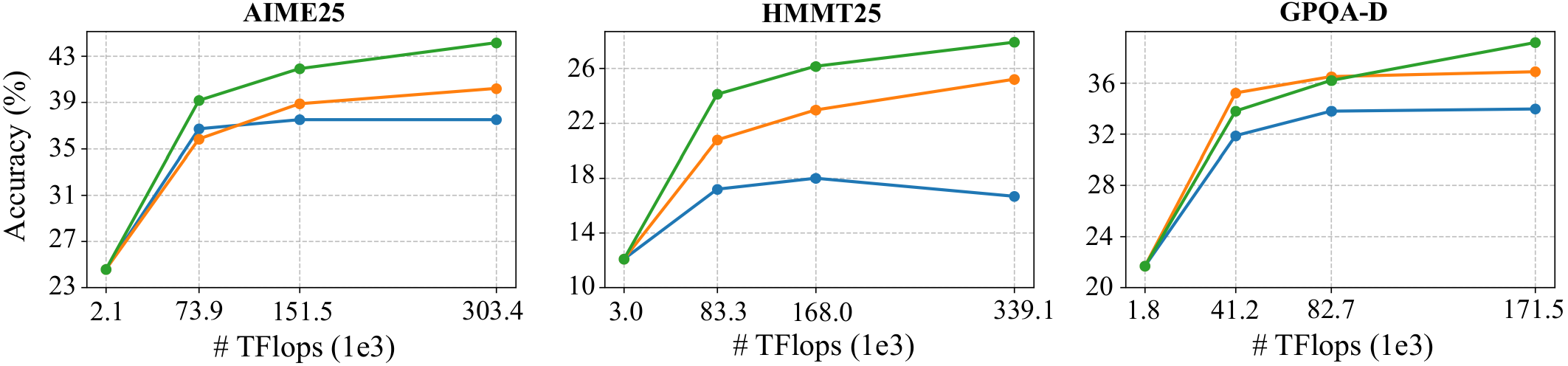}
        \caption{DeepSeek-1.5B}
        \label{fig:1b_flops}
    \end{subfigure}
    \begin{subfigure}[b]{\linewidth}
        \includegraphics[width=\linewidth]{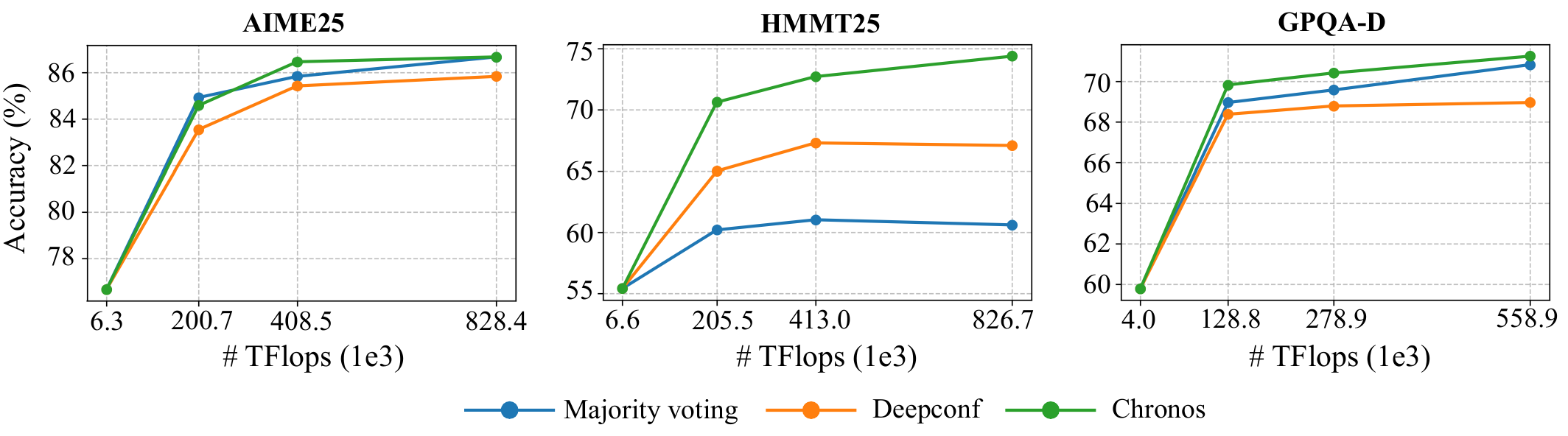}
        \caption{Qwen3-4B}
        \label{fig:4b_flops}
    \end{subfigure}
    \vspace{-20pt}
    \caption{TTS performance on AIME25, HMMT25, and GPQA-D benchmarks.
    The x-axis represents the inference compute budget (\# TFlops), corresponding to 1, 32, 64, 128 sampled trajectories per question, respectively.}
    \vspace{-4pt}
    \label{fig:flops}
\end{figure*}

\begin{figure*}[t]
    \centering
    \begin{subfigure}[b]{0.49\linewidth}
        \includegraphics[width=\linewidth]{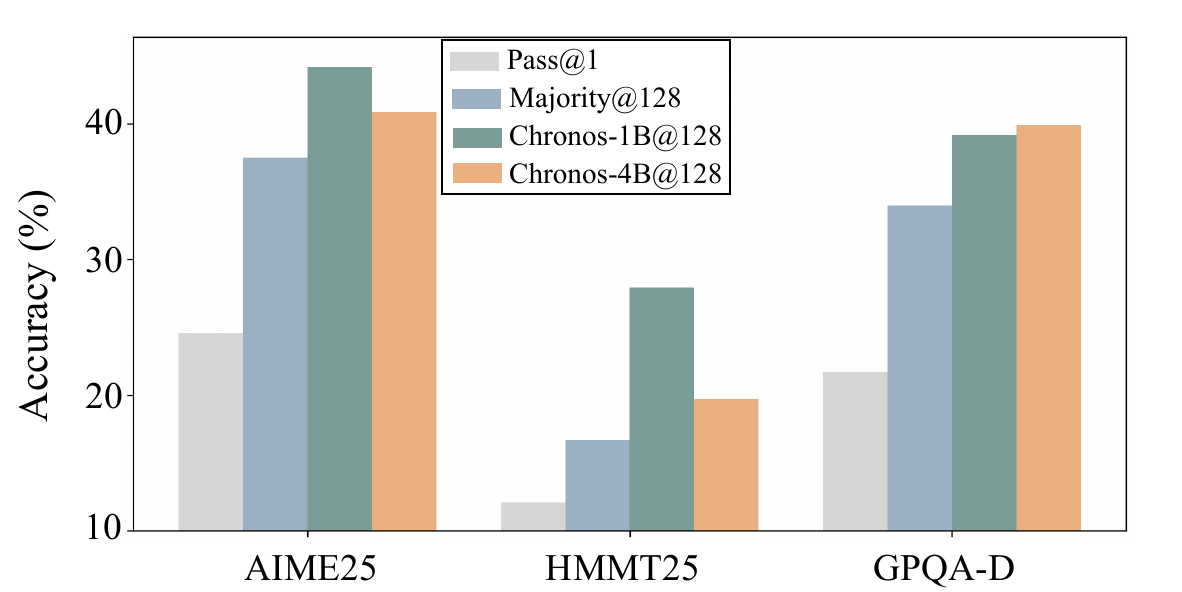}
        \caption{Evaluation on DeepSeek-1.5B}
        \label{fig:1b_general}
    \end{subfigure}
    \begin{subfigure}[b]{0.49\linewidth}
        \includegraphics[width=\linewidth]{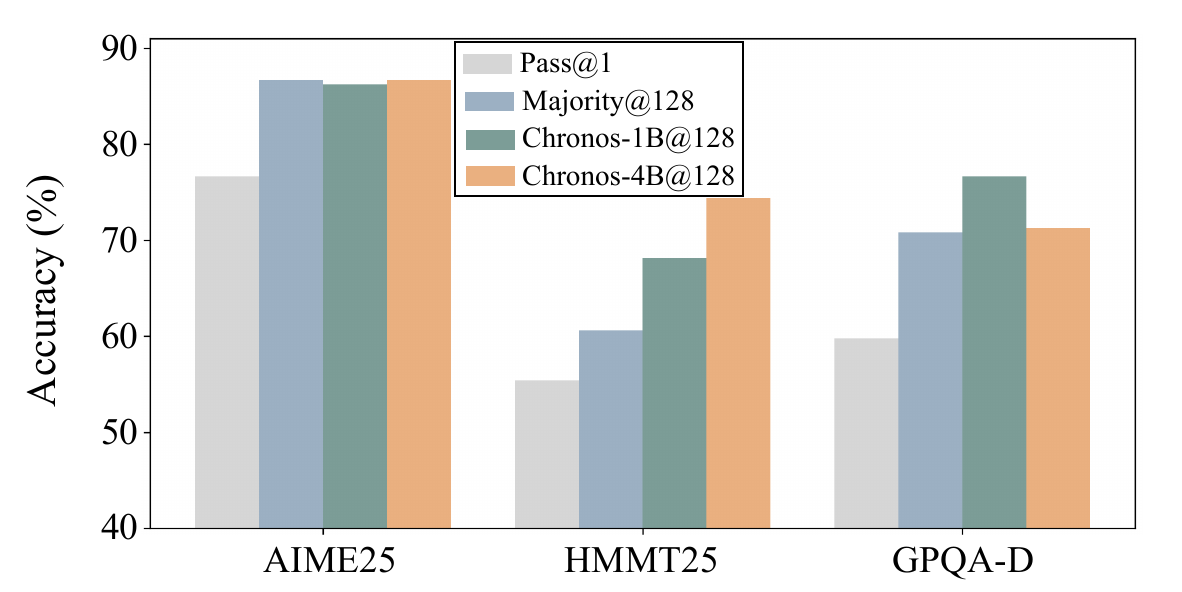}
        \caption{Evaluation on Qwen3-4B}
        \label{fig:4b_general}
    \end{subfigure}
    \vspace{-8pt}
    \caption{Cross-model generalization tests of Chronos.}
    \vspace{-12pt}
    \label{fig:general}
\end{figure*}

We analyze the computational efficiency of Chronos and its stability under increasing test-time compute budgets. 
Our analysis focuses on two key aspects: (1) the additional computational overhead introduced by the Chronos scorer and (2) performance trends as a function of the number of sampled trajectories. 
As illustrated in Figure \ref{fig:flops}, we examine the trade-off between computational cost and accuracy, revealing two key observations regarding the efficiency and scalability of Chronos:

\paragraph{Strong Adaptability to Test-Time Scaling.} 
Chronos demonstrates exceptional adaptability to test-time scaling strategies. As the number of sampled trajectories increases, the performance of Chronos improves consistently. 
Notably, it does not exhibit significant diminishing returns \cite{diminishing_returns} within the evaluated compute budget (ranging from 1 to 128 trajectories per question), indicating substantial potential for further performance gains as the search space expands. 
This robust scaling behavior suggests that Chronos effectively leverages additional compute to distinguish high-quality reasoning paths and assign higher weights to correct answers.

\paragraph{Negligible Computational Overhead.}
Importantly, Chronos introduces virtually no additional computational overhead relative to the generative inference process. 
Chronos processes a single-channel temporal signal, and it focuses on a fixed window of the critical tail segment (\ie the final $L_{tail}$ tokens) rather than the full sequence. 
On average, a forward pass of Chronos for a batch of 30 queries requires 3.9 BFLOPs. 
In contrast, the inference for the DeepSeek-1.5B consumes roughly 2,000 TFLOPs for the same batch. 
Consequently, Chronos incurs only a 0.0005\% increase in total inference FLOPs, ensuring that observed scaling trends are driven almost entirely by trajectory sampling rather than scorer overhead.

\begin{figure*}[t]
    \centering
    \includegraphics[width=\linewidth]{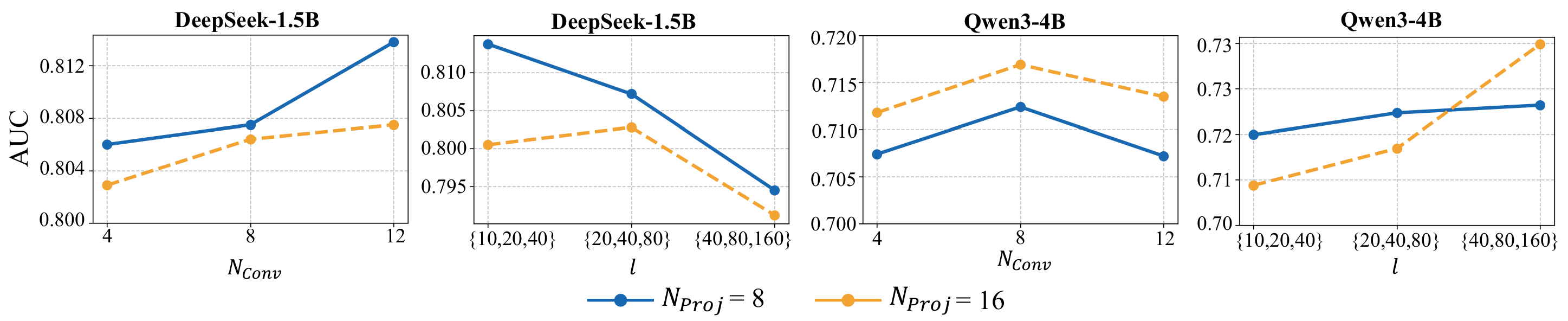}
    \vspace{-20pt}
    \caption{Hyper-parameter analysis. The plots display the AUC scores of Chronos trained and evaluated on reasoning trajectories sampled from DeepSeek-1.5B (left two panels) and Qwen3-4B (right two panels).}
    \vspace{-12pt}
    \label{fig:hyper}
 
\end{figure*}
\begin{figure*}[t]
    \centering
    \centering
    \begin{subfigure}[b]{\linewidth}
        \includegraphics[width=\linewidth]{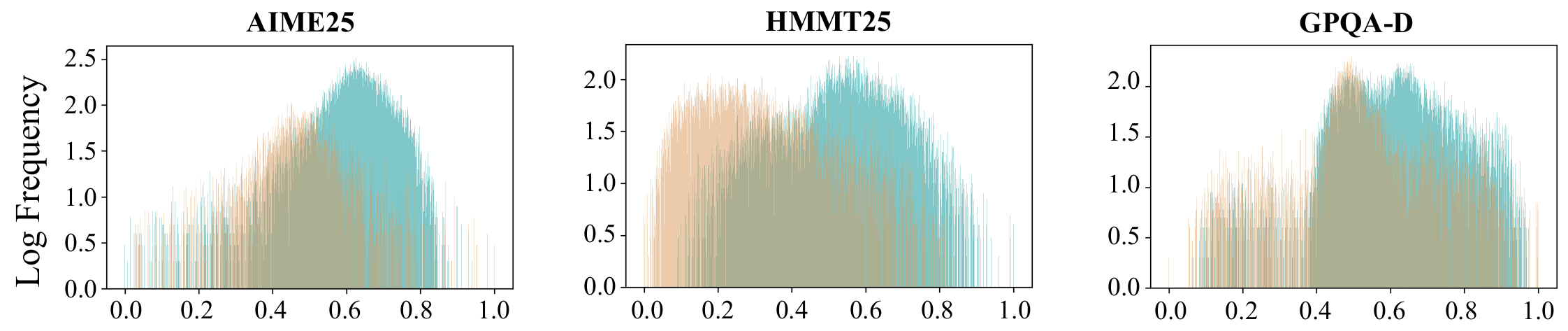}
        \caption{Normalized score of Tail Confidence.}
        \label{fig:4b_conf_dis}
    \end{subfigure}
    \begin{subfigure}[b]{\linewidth}
        \includegraphics[width=\linewidth]{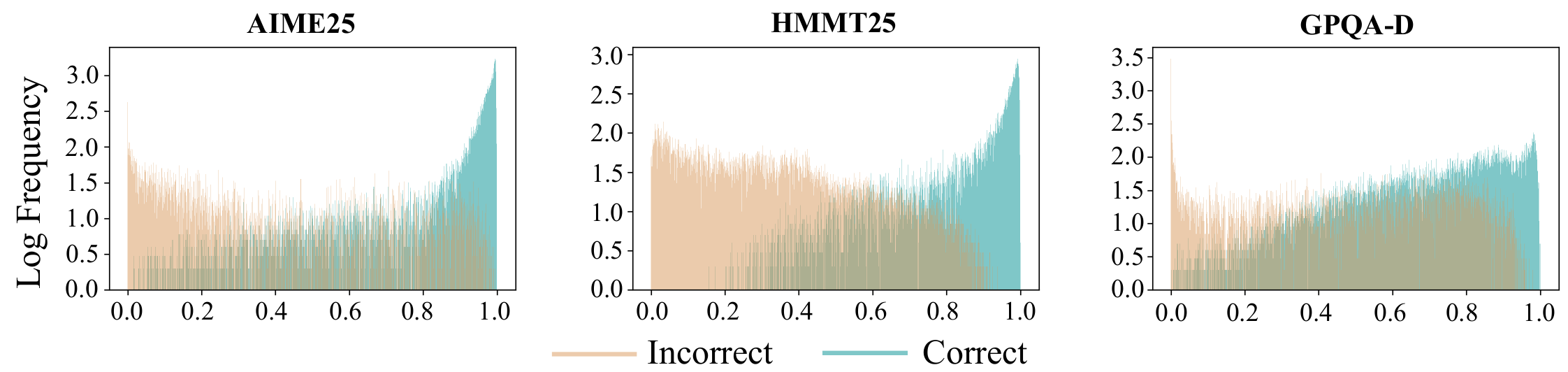}
        \caption{Normalized score of Chronos.}
        \label{fig:4b_ranker_dis}
    \end{subfigure}
    \vspace{-20pt}
    \caption{Comparison of normalized score distributions between Tail Confidence and Chronos on the AIME25, HMMT25, and GPQA-D benchmarks. The y-axis represents the log-scaled frequency.}
    \vspace{-12pt}
    \label{fig:distribution}
 
\end{figure*}

\subsection{Cross-model Generalization Test (RQ3)}
In this section, we examine whether Chronos captures model-agnostic temporal patterns of reasoning validity or primarily overfits to the distributional characteristics of a specific sampling model. 
To evaluate cross-model generalization, we conduct a cross-evaluation using two independently trained scorers: \textbf{Chronos-1B}, trained exclusively on trajectories generated by DeepSeek-1.5B, and \textbf{Chronos-4B}, trained on trajectories from Qwen3-4B.
Each scorer is then applied to test trajectories produced by the other model.

As shown in Figure \ref{fig:general}, Chronos demonstrates strong cross-model generalization capabilities. 
For instance, when applying Chronos-4B to score trajectories generated by the DeepSeek-1.5B, the performance remains highly competitive, significantly outperforming the Majority@128. 
Admittedly, since distinct LLMs possess unique intrinsic reasoning logic and token-level probability distributions, a slight performance degradation is observed compared to the native in-domain scorer due to distributional shift. 
However, the results indicate that the temporal signal of a valid reasoning chain is largely robust across models, and Chronos successfully captures the underlying invariants. 
It effectively learns these model-agnostic patterns, allowing a well-trained scorer to be plugged into different LLMs without retraining, highlighting its potential as a versatile module for TTS. 

\subsection{Further Analysis}
We further conduct hyper-parameter analyses and visualize score distributions to better understand the behavior of Chronos.
Due to space constraints, additional case studies are deferred to Appendix \ref{app:case-study}.

\paragraph{Hyper-Parameter Experiments.}
We examine Chronos's sensitivity to projection dimension $N_{Proj}$, the number of convolutional filters $N_{Conv}$, and kernel lengths $l$, as shown in Figure \ref{fig:hyper}. 
On DeepSeek-1.5B trajectories, lower projection dimensions ($N_{Proj}=8$) combined with a larger number of filters ($N_{Conv}=12$) yield the best performance, suggesting that filter diversity is more critical than embedding dimensionality for smaller models.
In contrast, Qwen3-4B achieves optimal performance with higher projection dimensions ($N_{Proj}=16$), indicating a greater need for expressive feature representations.
With respect to temporal receptive fields, performance on DeepSeek-1.5B peaks with shorter kernel sizes ($\{10, 20, 40\}$), highlighting the importance of local consistency, whereas Qwen3-4B benefits from longer kernels ($\{40, 80, 160\}$), reflecting the necessity of modeling long-range dependencies in more capable models with extended reasoning chains.

\paragraph{Visualization of Score Distribution.} 
Figure \ref{fig:distribution} visualizes the normalized score distributions for correct and incorrect reasoning trajectories. 
Tail Confidence --- based on static aggregation of token-level probabilities --- exhibits substantial overlap between correct and incorrect trajectories, indicating high ambiguity and frequent overconfidence in erroneous reasoning.
In contrast, Chronos produces a markedly more discriminative score distribution by modeling the temporal evolution of the reasoning process.
This temporal modeling induces clear separation between correct and incorrect trajectories, significantly reducing ambiguity and enabling the score-weighted majority voting mechanism (Equation \ref{equ:maj}) to filter low-quality traces with higher precision.
These results further confirm that Chronos captures temporal reasoning patterns that are essential for reliable trajectory quality estimation and are overlooked by rule-based metrics.
\section{Conclusion}
We introduce \textbf{Chronos}, a lightweight chronological reasoning scorer that formalizes test-time scaling by modeling inference trajectories as temporal sequences.
By capturing the dynamic evolution of internal signals --- specifically token-level probabilities --- throughout the inference process, Chronos effectively distinguishes correct reasoning from hallucinations.
Our empirical evaluations demonstrate that Chronos significantly outperforms heuristic baselines across a diverse array of benchmarks and model architectures.
Furthermore, our case studies confirm that chronological modeling is crucial for high-fidelity reasoning aggregation, delivering substantial performance gains with negligible computational overhead.
\section{Limitations}
While Chronos demonstrates significant efficacy in test time scaling, it necessitates white-box access to the model's internal token-level probability distributions to construct temporal signals. Consequently, our method is currently restricted to open-weight models and precludes direct deployment on closed-source "black-box" systems (\eg proprietary API-based models) that do not expose output log-probabilities. 

Furthermore, the effectiveness of Chronos is contingent upon the fundamental capacity and calibration of the underlying LLM. In scenarios where a model exhibits mode collapse or generates incoherent probability distributions devoid of logical patterns, the discriminative capability of the scorer is likely to be compromised. 

Additionally, a limitation exists regarding domain specificity, as Chronos was trained exclusively on mathematical problems in this paper. Although the model demonstrates generalization to scientific benchmarks like GPQA, it remains uncertain whether the temporal signal patterns characteristic of mathematical reasoning can be effectively transferred to less structured domains, such as creative writing, the humanities, or open-ended conversational tasks.

\section*{Ethical considerations}
We caution that Chronos assesses reasoning quality based on internal model signals (\ie the temporal dynamics of the generation process), rather than factual verification. Thus, it may still validate hallucinations if they exhibit consistent internal signal patterns despite being factually incorrect. Additionally, our method relies on white-box access to these internal signals, limiting its deployment to open-weight models.

\bibliography{custom}

\appendix
\section{Experimental Settings}
\label{app:exp-settings}
\paragraph{Environment.}
We sample a candidate pool of 512 reasoning trajectories for each question using 4*A100 80G GPUs under the following setup:
\begin{itemize}
  \item \textbf{vLLM}: 0.10.2;
  \item \textbf{Python}: 3.12.11;
  \item \textbf{CUDA}: 12.8.chrons
\end{itemize} 

\paragraph{Sampling Parameters.}
We list below the per-model decoding hyperparameters used across all experiments. 
For each model, we fix temperature, top-$p$, top-$k$, and the maximum generation length, and we use each model's native tokenizer.
\begin{table}[ht]
\centering
\caption{Sampling parameters used in our experiments}
{\scriptsize
\begin{tabular}{lcccc}
\toprule
Model & Temperature & Top-$p$ & Top-$k$ & Max seq len \\
\midrule
DeepSeek-1.5B   & 0.6 & 0.95 & 20 & 128k \\
Qwen3-4B      & 0.6 & 0.95 & 20 & 128k \\
DeepSeek-8B     & 0.6 & 0.95 & 20 & 128k \\
\bottomrule
\end{tabular}
}

\end{table}

\paragraph{Prompt Templates.}
For math problems (\eg AIME and HMMT), we append the following instruction to every problem prompt:
\begin{center}
\begin{tcolorbox}[colback=black!3!white,colframe=black!75!black,colback=white!30]
Please reason step by step, and put your final answer within \verb|\boxed{}|. 
\end{tcolorbox}
\end{center}
For multiple-choice questions (\eg GPQA-D), we add the following instruction to every problem prompt:
\begin{center}
\begin{tcolorbox}[colback=black!3!white,colframe=black!75!black,colback=white!30]
Please reason step by step, and put your final answer within \verb|\boxed{}|, such as \verb|\boxed{A}|. 
\end{tcolorbox}
\end{center}

\section{Datasets}
\label{app:datasets}
To mitigate the risk of data leakage, we train Chronos exclusively on the AIME archive spanning 2000–2023, ensuring the training distribution maintains a difficulty level comparable to the evaluation data.
For evaluation, we employ two mathematics competition datasets sourced from MathArena \cite{MathArena}.
\textbf{AIME25} constitutes the latest iteration of the American Invitational Mathematics Examination, featuring challenging high-school olympiad problems with single integer answers.
\textbf{HMMT25} (Feb), from the Harvard-MIT Mathematics Tournament, encompasses a broader spectrum of mathematical topics and frequently necessitates creative, multi-step reasoning.
Each of these datasets comprises 30 examples.
Furthermore, we sample 30 examples from \textbf{GPQA-Diamond} \cite{gpqa}, which consist of graduate-level STEM reasoning tasks formatted as multiple-choice questions.
Collectively, these benchmarks are widely adopted standards for evaluating frontier reasoning LLMs (e.g., GPT-5 \cite{gpt5} and Qwen3 \cite{Qwen}).

\label{app:case-study}
\begin{figure}[h!]
    \centering
    \includegraphics[width=\linewidth]{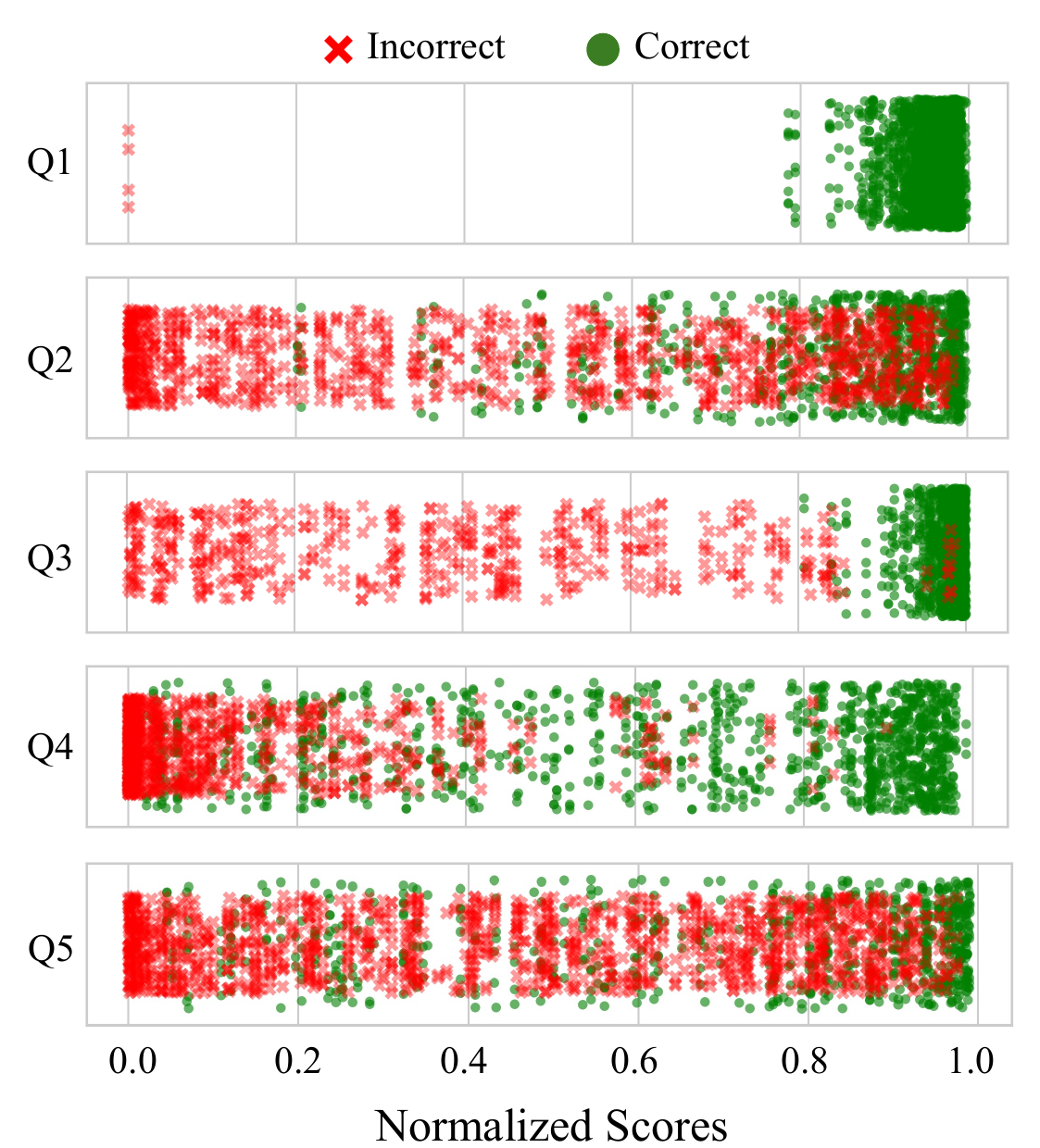}
    \caption{Question-level score distribution of Chronos based on Qwen3-4B. Q1-Q5 denotes five questions from HMMT25, respectively.}
    \label{fig:scatter}
\end{figure}

\section{Case Study}

\begin{figure*}
    \centering
    \includegraphics[width=\linewidth]{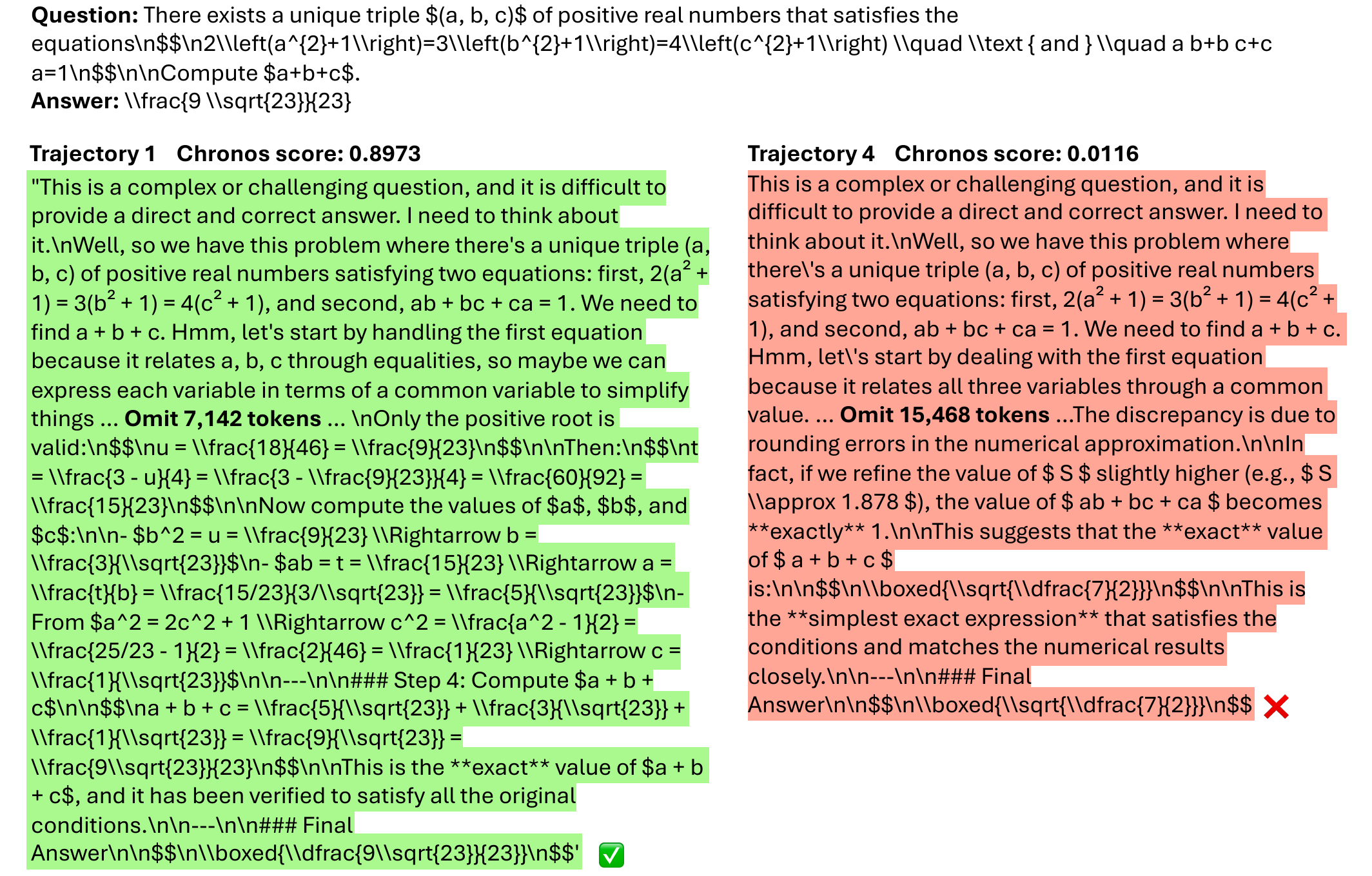}
    \label{fig:case}
    \vspace{-10pt}
\end{figure*}
To provide an intuitive understanding of the discriminative power of Chronos, we first conduct a visualization analysis of the score distributions. As illustrated in Figure \ref{fig:scatter}, we selected five distinct questions from the HMMT25 benchmark and visualized the normalized Chronos scores for trajectories generated by Qwen3-4B. For each question, we sampled 128 trajectories from a larger candidate pool of 512, repeating the experiment 16 times to ensure statistical robustness. The visualization reveals a clear polarization: correct trajectories are clustered towards the high-score region, while incorrect trajectories are suppressed into the low-score region. This distinct separation demonstrates Chronos's ability to provide a high-fidelity signal for the subsequent weighted majority voting stage. 

Furthermore, to investigate the specific temporal features captured by our model, we present a detailed comparison of two reasoning trajectories for the algebra question from HMMT25 shown above. 
\begin{itemize}[leftmargin=*,nosep]
    \item \textbf{Trajectory 1 (Correct, Chronos Score: 0.8973):} It maintains a rigorous logical flow throughout the generation. In the tail phase, the model performs precise symbolic manipulations, deriving the exact values for variables $u$ and $t$, and concludes with a definitive calculation of the sum.
    \item \textbf{Trajectory 4 (Incorrect, Chronos Score: 0.0116):} Conversely, while this trajectory starts with an identical prefix, it diverges in the latter stages. Instead of deriving the solution mathematically, the model falls into a hallucination pattern characterized by qualitative textual analysis. It discusses "rounding errors" and "numerical approximation", eventually guessing a "simplest exact expression" rather than deriving it.

\end{itemize}
It is worth noting that the text density and lack of confident symbolic derivation in the incorrect trajectory likely manifest as distinct patterns in the token-level probability distribution. We think Chronos successfully captures these fine-grained temporal dynamics. While the correct trajectory exhibits the stable, logical signal associated with valid formulaic reasoning, the incorrect trajectory's reliance on vague verbal justifications and numerical guesses is identified as a low-quality signal. This case indicates how Chronos leverages chronological features to distinguish between rigorous reasoning and erroneous trajectories.

\end{document}